%% file: main.tex
\PassOptionsToPackage{dvipsnames}{xcolor}
\documentclass{article}

\usepackage{microtype}
\usepackage{graphicx}
\usepackage{subfigure}
\usepackage{booktabs} 

\usepackage{hyperref}

\usepackage[accepted]{icml2025}

\usepackage{amsmath}
\usepackage{amssymb}
\usepackage{mathtools}
\usepackage{amsthm}

\usepackage[capitalize,noabbrev]{cleveref}

\theoremstyle{plain}

\theoremstyle{definition}

\theoremstyle{remark}

\usepackage[textsize=tiny]{todonotes}

\usepackage{epsfig}
\usepackage{graphicx}
\usepackage{amsmath}
\usepackage{amssymb}
\usepackage{booktabs}
\usepackage{multirow}
\usepackage{tabularx}
\usepackage{enumitem}

\usepackage{titletoc}
\usepackage{bbding}
\usepackage[marginal]{footmisc}

\usepackage[dvipsnames]{xcolor}
\usepackage{colortbl}
\usepackage{caption}
\usepackage{makecell}
\usepackage{amssymb}
\usepackage{pifont}
\definecolor{mycyan}{cmyk}{.1,0,0,0}
\definecolor{mygray}{gray}{.95}
\definecolor{mypink}{rgb}{.99,.91,.95}
\newcommand{\cmark}{\ding{51}}%
\newcommand{\xmarkg}{\textcolor{lightgray}{\ding{55}}}%
\newcommand{\mypara}[1]{\vspace{1mm}\noindent\textbf{#1}}
\usepackage[capitalize]{cleveref}
\newcommand{\name}{S2-Track}
\definecolor{F7E0D5}{RGB}{245,240,255}

\newcommand{\pub}[1]{\color{gray}{\scriptsize{[{#1}]}}}

\newcommand\decoder[0]{{\color{Blue}Uncertainty-aware Probabilistic Decoder}}
\newcommand\init[0]{{\color{Orange}2D-Prompted Query Initialization}}
\newcommand\denoise[0]{{\color{Green}Hierarchical Query Denoising}}

\newcommand\UPD[0]{{\color{Blue}UPD}}
\newcommand\PQI[0]{{\color{Orange}PQI}}
\newcommand\HQD[0]{{\color{Green}HQD}}

\icmltitlerunning{Submission and Formatting Instructions for ICML 2025}

\begin{document}

\twocolumn[
\icmltitle{\name: A Simple yet Strong Approach for End-to-End \\ 3D Multi-Object Tracking}



\icmlsetsymbol{equal}{*}

\begin{icmlauthorlist}

\icmlauthor{Tao Tang$^{\ddagger}$}{equal,sch}
\icmlauthor{Lijun Zhou}{equal,comp}
\icmlauthor{Pengkun Hao}{comp}
\icmlauthor{Zihang He}{comp}
\icmlauthor{Kalok Ho}{comp}
\icmlauthor{Shuo Gu}{comp}
\icmlauthor{Zhihui Hao}{comp}
\icmlauthor{Haiyang Sun}{comp}
\icmlauthor{Kun Zhan}{comp}
\icmlauthor{Peng Jia}{comp}
\icmlauthor{Xianpeng Lang}{comp}
\icmlauthor{Xiaodan Liang$^{\dagger}$}{sch}

\end{icmlauthorlist}

\icmlaffiliation{sch}{Shenzhen Campus of Sun Yat-sen University}
\icmlaffiliation{comp}{Li Auto Inc}

\icmlcorrespondingauthor{Xiaodan Liang}{liangxd9@mail.sysu.edu.cn}

\icmlkeywords{Machine Learning, ICML}

\vskip 0.3in
]



\printAffiliationsAndNotice{\icmlEqualContribution, $^{\ddagger}$ Work done during an internship at Li Auto Inc.}

\input{sec/0_abstract}
\input{sec/1_intro}
\input{sec/2_related}

\input{sec/3_method}
\input{sec/4_exp}
\input{sec/5_conclusion}

\section*{Acknowledgments}
This work is supported by Scientific Research Innovation Capability Support Project for Young Faculty (No.ZYGXQNJSKYCXNLZCXM-I28), National Natural Science Foundation of China (NSFC) under Grants No.62476293, Shenzhen Science and Technology Program No.GJHZ20220913142600001, Nansha Key R\&D Program under Grant No.2022ZD014, and General Embodied AI Center of Sun Yat-sen University.    

\section*{Impact Statement}
This paper presents a strong yet simple tracker, S2-Track, for 3D multiple object tracking. Since the tracking explored in this paper is for generic objects and does not pertain to specific human recognition, so we do not see potential privacy-related issues.
At present, we primarily focus on the image representation of scenes. However, our framework can be expanded to incorporate other sensors, such as a fusion of LiDAR and cameras. We leave the extension of our method towards building such systems for future work. We hope that our work can inspire more
future research in this field.

\nocite{langley00}

\bibliography{main}
\bibliographystyle{icml2025}

\newpage
\appendix
\onecolumn

\input{sec/supp}

\end{document}

%% file: sec/0_abstract.tex
\begin{abstract}
  3D multiple object tracking (MOT) plays a crucial role in autonomous driving perception. 
  Recent end-to-end query-based trackers simultaneously detect and track objects, which have shown promising potential for the 3D MOT task.
  However, existing methods are still in the early stages of development and lack systematic improvements,
  failing to track objects in certain complex scenarios, like occlusions and the small size of target object's situations.
  In this paper, we first summarize the current end-to-end 3D MOT framework by decomposing it into three constituent parts: query initialization, query propagation, and query matching. Then we propose corresponding improvements, which lead to a strong yet simple tracker: \name. 
  Specifically, for query initialization, we present 2D-Prompted Query Initialization, which leverages predicted 2D object and depth information to prompt an initial estimate of the object's 3D location.
  For query propagation, we introduce an Uncertainty-aware Probabilistic Decoder to capture the uncertainty of complex environment in object prediction with probabilistic attention. 
  For query matching, we propose a Hierarchical Query Denoising strategy 
  to enhance training robustness and convergence.
  As a result, our \name{} achieves state-of-the-art performance on nuScenes benchmark, i.e., 66.3\% AMOTA on test split, surpassing the previous best end-to-end solution by a significant margin of 8.9\% AMOTA. 
  We achieve 1st place on the \href{https://www.nuscenes.org/tracking?externalData=no&mapData=no&modalities=Camera}{nuScenes tracking task leaderboard}.

\end{abstract}

%% file: sec/1_intro.tex
\section{Introduction}

3D multiple object tracking (MOT)~\cite{li2023dqtrack, doll2023star, pang2023PFtrack, qing2023dort,yang2022QTtrack, li2023poly, wang2023camo, ding2024adatrack} is an essential component for the perception of autonomous driving systems. The ability to accurately and robustly track objects in dynamic environments is crucial for ensuring smooth and safe navigation and reasonable decision-making. 
Traditional 3D MOT methods~\cite{yang2022QTtrack, chaabane2021deft, zhou2020trackingL2, yin2021centerL2,pang2022simpletrackIOU, weng20203dKalman, wojke2017simpleKalman, guo2024cyclic} rely on detector outcomes followed by a post-processing module like data association and trajectory filtering, leading to a complex pipeline.
To avoid human-crafted heuristic design in detection-based trackers, recent advancements in end-to-end query-based approaches have shown impressive potential in addressing the 3D MOT task by simultaneously detecting and tracking objects~\cite{zeng2022motr, zhang2022mutr3d, doll2023star, li2023dqtrack, pang2023PFtrack, ding2024adatrack}. These methods have demonstrated promising results in terms of tracking performance and efficiency.
However, current end-to-end trackers are still in the early stages of development and can not effectively handle the various complex driving scenarios with their naive solution. 

\input{latex/fig/teaser}

In driving scenarios, the environment could be highly complex, often when driving in cities, with numerous objects such as vehicles and pedestrians 
interleaving across the scene, and exhibiting substantial variations in their motion patterns.
Furthermore, tracked objects often cover a wide spatial tracking range and a long temporal tracking sequence.
As a result, occlusion situations and the small size of target objects, frequently occur, which usually leads to some undetected or occluded objects losing track.
These factors present significant challenges to current vanilla end-to-end query-based approaches for achieving accurate and robust 3D MOT.
As shown in \cref{fig:teaser} (a), the previous state-of-the-art end-to-end tracker, PF-Track~\cite{pang2023PFtrack}, fails to track objects in challenging scenarios.

In this paper, we aim to comprehensively enhance the existing end-to-end 3D MOT framework to achieve robust and accurate tracking results in complex driving environments.
As illustrated in \cref{fig:teaser} (b), we first delve into the current query-based framework and decompose it into three constituent parts: query initialization, query propagation, and query matching. Then we propose corresponding improvements, which lead to a \textbf{S}trong yet \textbf{S}imple tracker: \name.  
Firstly, for \textit{query initialization}, we present the \init{} module, which leverages predicted 2D object location and depth information to 
enhance the accuracy of initial object localization, leading to more reliable tracking results.
Secondly, for \textit{query propagation}, we introduce an \decoder{} to capture and model the uncertainty of complex environments during object prediction. Specifically, we model attention scores as Gaussian distributions instead of deterministic outputs, to quantify the predictive uncertainty.
Moreover, for \textit{query matching}, we propose an \denoise{} strategy to further improve the training process. During the training stage, we add noises to ground-truth bounding boxes to form noised queries and selectively denoise queries based on their noised levels, enhancing robustness and convergence.
Experimental results on the nuScenes benchmark demonstrate the effectiveness of our \name{} framework. It achieves state-of-the-art performance with an impressive 66.3\% AMOTA on the test split, surpassing the previous best end-to-end solution by a significant margin of 8.9\% AMOTA. 
These results highlight our simple yet non-trivial improvements and showcase the potential of our framework in advancing the field of autonomous driving perception.

To summarize, our contributions are as follows:
\begin{itemize}
\item We delve into the current end-to-end 3D MOT framework and decompose it into three constitute modules, and propose a stronger yet simple framework, \name{}, which enhances each module comprehensively.
\item We propose three well-designed modules, 2D-Prompted
Query Initialization, Uncertainty-aware Probabilistic Decoder, and Hierarchical Query Denoising to enhance the previous pipeline from multiple aspects, leading to improved tracking performance.
\item We demonstrate the effectiveness of our \name{} framework quantitatively and qualitatively through extensive experiments on nuScenes benchmark and achieve leading performance with a remarkable 66.3\% AMOTA.
\end{itemize}


%% file: latex/fig/teaser.tex
\begin{figure}[t]
  \centering
  \includegraphics[width=1\linewidth]{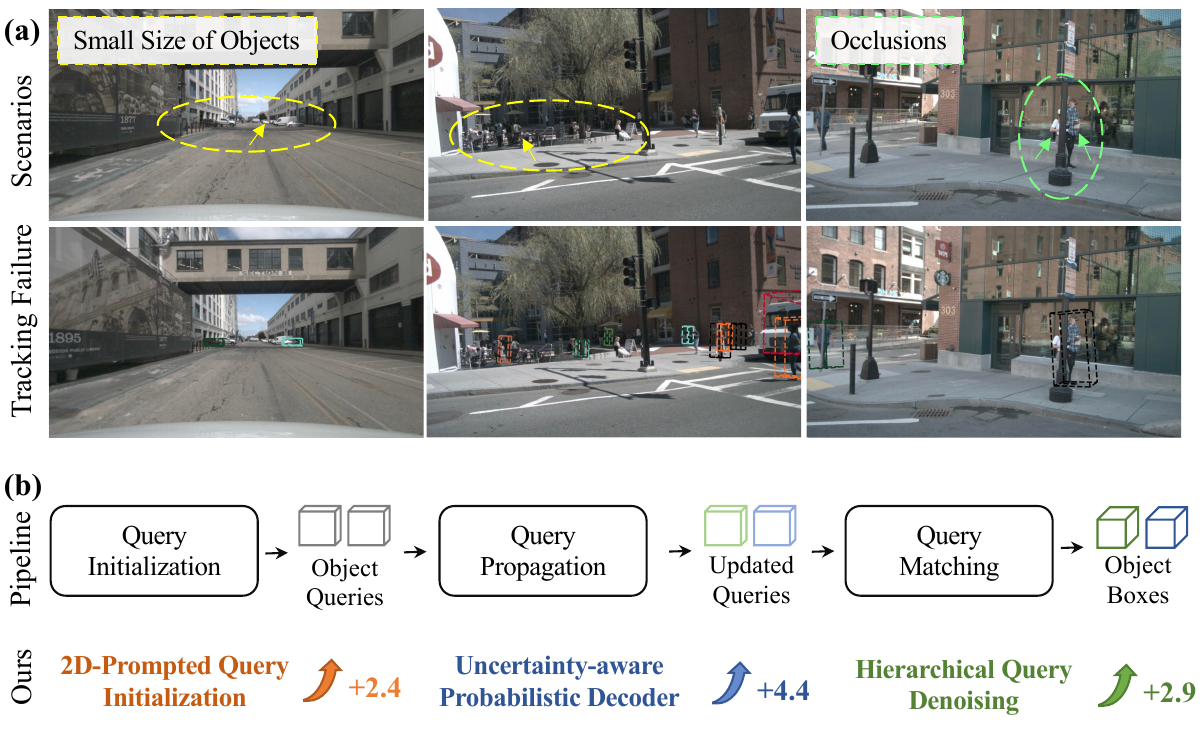}
  \caption{\textbf{(a)} In complex driving scenarios, there are various challenge factors, e.g., the occlusions and small size of target objects, which present significant challenges to achieving accurate tracking. The previous state-of-the-art end-to-end tracker, PF-Track~\cite{pang2023PFtrack}, fails to track objects in certain complex scenarios. \textbf{(b)} Our \name{} proposes three simple yet strong modules to enhance baseline comprehensively, leading to improved tracking performance.
  }
  \label{fig:teaser}
\end{figure}

%% file: sec/2_related.tex
\section{Related Work}
\mypara{Tracking by Detection.}
Multi-object tracking (MOT) in 3D scenes takes multi-view images from surrounding cameras or LiDAR point clouds to track multiple objects across frames~\cite{ marinello2022triplettrack,
wang2023StreamPETR, qing2023dort, cc3dt,yang2022QTtrack, 
li2023poly, wang2023camo, sadjadpour2023shasta, guo2024cyclic}.
Taking advances in 3D object detection ~\cite{huang2021bevdet, li2023bevdepth, li2022bevformer, liang2022bevfusion, liu2023bevfusion, liu2023petrv2, lin2023sparse4d, wang2022detr3d}, most 3D MOT methods follow the \textit{tracking by detection} paradigm~\cite{yang2022QTtrack, chaabane2021deft, zhou2020trackingL2, zhang2022bytetrack, yin2021centerL2,pang2022simpletrackIOU}, where tracking is treated as a post-processing step after object detection. Take the detected objects at each frame, traditional 3D MOT usually uses motion models, e.g., Kalman filter~\cite{weng20203dKalman, wojke2017simpleKalman}, to predict the status of corresponding trajectory and associate the candidate detections using 3D IoU~\cite{pang2022simpletrackIOU, weng20203dKalman} or L2 distance~\cite{zhou2020trackingL2, yin2021centerL2}. 

\input{latex/fig/main_pipline}
\mypara{Tracking with Query.}
To overcome the independent nature of the detection and tracking and to implicitly solve the association between frames, the recent \textit{tracking with query} paradigm models the tracking process with transformer queries~\cite{zeng2022motr, zhang2022mutr3d, doll2023star, li2023dqtrack, pang2023PFtrack, ding2024adatrack}.
MUTR3D~\cite{zhang2022mutr3d} extends the object detection method DETR3D~\cite{wang2022detr3d} for tracking by utilizing a 3D track query to jointly model object features across timestamps and multi-view.
STAR-TRACK~\cite{doll2023star} proposes a latent motion model to account for the effects of ego and object motion on the latent appearance representation.
DQTrack~\cite{li2023dqtrack} separates object and trajectory representation using decoupled queries, allowing more accurate end-to-end 3D tracking. 
PF-Track~\cite{pang2023PFtrack} extends the temporal horizon to provide a strong spatio-temporal object representation.
Although these methods achieved impressive performance, when applied to complex scenarios, especially the occlusions and the small target objects, the tracking performance becomes unsatisfactory. 
In this work, we delve into the current end-to-end 3D MOT framework and decompose it into three modules, and further propose corresponding
improvements to enhance each module comprehensively.
As a result, these improvements lead to a strong yet simple framework, \name{}, and consequently achieve leading performance.

%% file: latex/fig/main_pipline.tex
\begin{figure*}[t]
  \centering
  \includegraphics[width=1\linewidth]{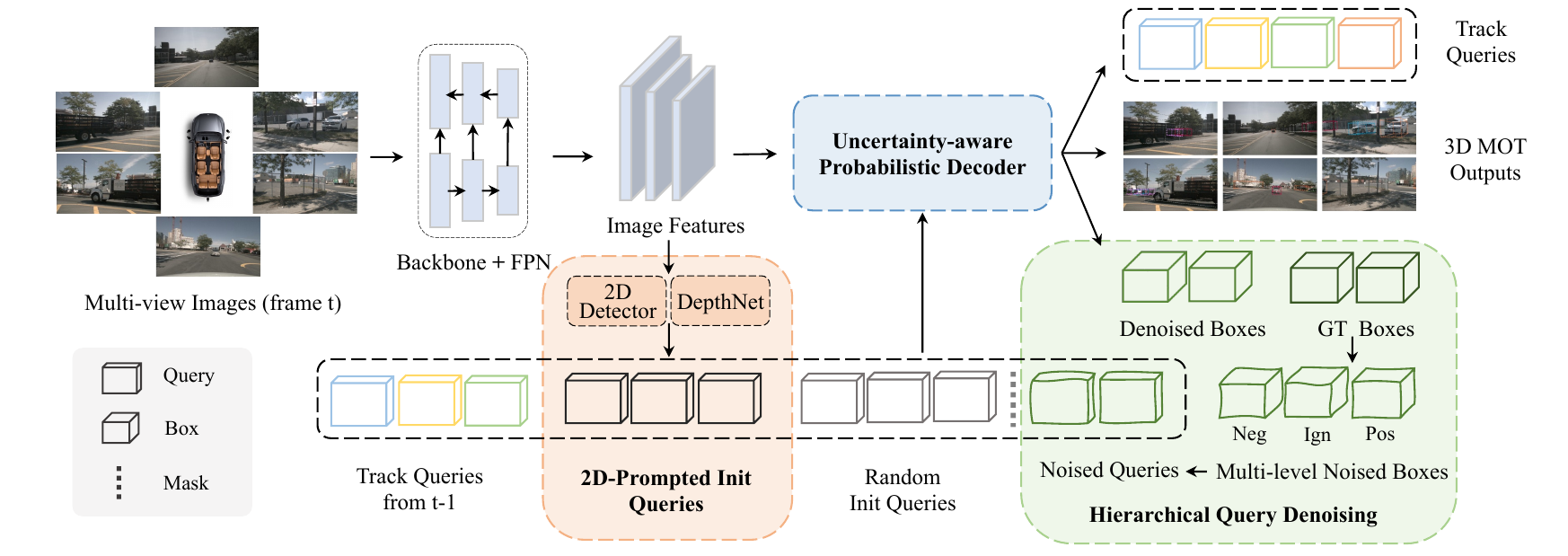}
  \caption{\textbf{\name{} framework}. 
  The proposed 2D-Prompted Query Initialization (\PQI{}), Uncertainty-aware Probabilistic Decoder (\UPD{}), and Hierarchical Query Denoising (\HQD{})
  are incorporated together to improve tracking performance. Neg: negative, Ign: ignore, Pos: positive, Mask: separate the normal queries and the denoising part to prevent information leakage.
  }
  \label{fig:main_pipline}
 \vspace{-10pt}
\end{figure*}

%% file: sec/3_method.tex
\section{\name{}}

In this section, we introduce our \name{} in detail. We first give a brief problem definition and an overview of the framework in \cref{fig:main_pipline}. Then, we clarify our key contributions:
2D-Prompted Query Initialization, Uncertainty-aware Probabilistic Decoder, and Hierarchical Query Denoising.

\subsection{Preliminaries}
\label{subsec:problem}

\mypara{Overview.} At each timestamp $t$, given $c$ images from surrounding cameras, the tracking objective is to estimate a set of bounding boxes $\mathbf{b}_t^{id} \in \textbf{B}_t$ with consistent $id$ across frames.
Under the \textit{tracking with query} paradigm, an overview of proposed \name{} is presented in \cref{fig:main_pipline}, which is conceptually simple: encoder and transformer decoder are adopted to encode input images and decode 3D MOT outputs with queries. From the perspective of the query, it can be divided into three stages: query initialization, query propagation with image features, and query matching with ground truth.

\mypara{Queries initialization.}
Specifically, following previous studies~\cite{pang2023PFtrack, ding2024adatrack}, our \name{} utilizes a set of object queries to tackle multi-object tracking from multi-view images. 
Each query $\textbf{q}_t^i\in \textbf{Q}_t$ represents a unique 3D object with a feature vector $\textbf{f}_t^i$ and a 3D location $\textbf{c}_t^i$, i.e., $\textbf{q}^i_t = \{\textbf{f}_t^i, \textbf{c}_t^i\}$.
The queries are randomly initialized as learnable embeddings, denoted as $\textbf{Q}_{init}$.
Here, we propose \init{} (orange module) to improve the query initialization with predicted 2D object location and depth information. 

\mypara{Queries propagation.}
During training, the object is tracked by updating its unique query. Specifically, the object queries $\textbf{Q}_t=\{\textbf{q}^i_t\}$ are propagated from the previous frame $t-1$ (colored squares) and numerous initial queries (gray squares in \cref{fig:main_pipline}):
\begin{equation}
    \textbf{Q}_t\leftarrow \mathbf{Prop}(\textbf{Q}_{t-1}, \textbf{Q}_{init}).
    \label{eq:prop}
\end{equation}
The queries from the previous frame represent tracked instances, while numerous initial queries aim to discover new objects. 
Here, we introduce an \decoder{} (blue module) to model and capture the uncertainty of complex environments in object prediction.

\mypara{Queries matching.}
Then, to predict 3D bounding boxes, decoder-only transformer architectures such as DETR3D~\cite{wang2022detr3d} and PETR~\cite{liu2023petrv2}, are utilized to decode image features $\textbf{F}_t$ with object queries:
\begin{equation}
   \textbf{B}_t,\textbf{Q}_t\leftarrow \mathbf{Decoder}(\textbf{F}_t,\textbf{Q}_t),
   \label{eq:det}
\end{equation}
where $\textbf{B}_t$ and $\textbf{Q}_t$ are the detected 3D bounding boxes and updated query features respectively.
Here, we present an \denoise{} strategy (green module) to enhance the model robustness and convergence.

In the following sections, we give detailed elaboration.

\subsection{\init{}} 
\label{subsec:initial}
In \textit{tracking with query} frameworks, high-quality initial queries are crucial for achieving rapid convergence and improving tracking precision. 
This becomes particularly important when dealing with complex driving scenarios, such as occlusion and small-sized target objects. In previous methods that solely rely on random query initialization, the lack of reliable initial queries often results in some undetected or occluded objects losing track.
To address this, we propose 2D-\textbf{P}rompted \textbf{Q}uery \textbf{I}nitialization (\PQI{}) module which enhances the initialization of queries using learned certain priors obtained from network training.
Specifically, after utilizing the shared image backbone and the feature pyramid network (FPN) layers to extract image features from each camera, we introduce additional auxiliary tasks, i.e., the 2D detection and the depth prediction, as follows:
\begin{align}
\textbf{B}_{t}^{2d}, \textbf{D}_{t} \leftarrow \text{Networks}_\text{{auxiliary}}(\textbf{F}_t),
 \label{eq:depth}
\end{align}
where $\textbf{B}_{t}^{2d}, \textbf{D}_{t}$ denotes the 2D bounding boxes and the depth respectively. The 2D detection head follows YOLOX~\cite{yolox2021}.
The depth network combines multiple residual blocks 
and is supervised with the projected LiDAR points. The optimization objectives of the two auxiliary tasks are:
\begin{equation}
\mathcal{L}_{\PQI{}} = \mathcal{L}_{Det}^{2D} +  \mathcal{L}_{Depth}.
\label{eq:losscei}
\end{equation}

Then, we estimate 3D location $\textbf{C}_t=\{\textbf{c}^i_t\}$ through the coordinate transformation:
$
    \textbf{C}_t = T_{cam}^{lidar} K^{-1} \textbf{D}_{t} [u,v,1]^T,
$
where $T_{cam}^{lidar}$ and $K^{-1}$ represent the transformation matrix from the camera coordinate system to the lidar coordinate system and the camera's intrinsic parameters respectively, and $(u,v)$ denotes the center location of the 2D boxes. 
Then we initialize our object queries with the preliminary 3D location, denoted as $\textbf{Q}_\text{PQI-init}$ (\textcolor[RGB]{37, 37, 37}{dark gray squares} in \cref{fig:main_pipline}).
We also retain the random initialization $\textbf{Q}_\text{init}$ to 
explore missing objects.
To summarize, the query initialization and propagation process in \cref{eq:prop} can be improved as follows:
\begin{equation}
\textbf{Q}_t\leftarrow \mathbf{Prop}(\textbf{Q}_{t-1}, \textbf{Q}_\text{\PQI{}-init}, \textbf{Q}_\text{init}).
\label{eq:prop_our}
\end{equation}
We also provide qualitative results in Fig. 6 of Appendix, which shows the initial queries generated by our PQI module are accurately positioned within the regions of interest for the objects, resulting in more accurate tracking results.

\input{latex/fig/attention}
\subsection{\decoder{}}
\label{subsec:decoder}
In complex driving scenarios, the trajectories of multiple targets exhibit substantial variations in their temporal duration and sequence.
Furthermore, the target objects themselves can vary in size, ranging from large trucks to small children. 
These diversities present significant challenges for current end-to-end 3D MOT methods, leading to uncertainties. The uncertainty issue refers to neural networks that do not deliver certainty estimates or suffer from underconfidence, as current methods struggle to capture the noise and variations inherent in the input data.
Although the uncertainty issue has been recognized in certain fields~\cite{gawlikowski2021survey, subedar2019baydnn, wang2021DUGM}, e.g., action recognition~\cite{guo2022UGPT} and camouflaged object detection~\cite{yang2021UGTR}, it has not been discussed or explored in the context of 3D MOT.
Moreover, due to the complex driving scenarios and the unique challenge of tracking tasks, previous solutions for specific domains cannot be directly applied here. 

To address this limitation, inspired by previous uncertainty quantify works~\cite{guo2022UGPT, pei2022HSA, blundell2015weight}, we introduce \textbf{U}ncertainty-aware \textbf{P}robabilistic \textbf{D}ecoder (\UPD{}) for 3D MOT. 
Current end-to-end methods employ decoders with conventional transformers using deterministic attention mechanisms, which compute determinate attention $\alpha$ between queries ($Q$) and keys ($K$) as $\alpha = \frac{Q \cdot K}{\sqrt{d}}$, where $d$ is the dimension of queries and keys. The deterministic attention limits the ability to quantify uncertainty in predictions effectively.
Instead, our UPD utilizes the probabilistic attention computation, which assumes attention $\alpha$ follows a Gaussian distribution: $\alpha_{ij} \sim \mathcal{N}(\mu_{ij}, \sigma_{ij})$. Through the reparameterization trick~\cite{kingma2015variational}:
$
 \alpha_{ij} = \mu_{ij} + \sigma_{ij} \epsilon,  \epsilon \sim \mathcal{N}(0, 1).
$
As illustrated in \cref{fig:attention}, a multi-layer perception is adopted to fit the mean and standard deviation with $q_{i}, k_{j}$ as input: $\mu_{ij}, \sigma_{ij}^2 = MLP(q_{i}, k_{j})$. 
Thus, we introduce the uncertainty-aware probabilistic parameters $\mu_{ij}$ and $\sigma_{ij}$ into the decoder, 
allowing uncertainty adaptation in the training process.
Practically, we utilize the negative log-likelihood loss to constrain the probabilistic attention as:
\begin{equation}
\mathcal{L}_{\UPD{}} = \sum_{i,j} \log(\frac{1}{\sqrt{2\pi \sigma_{ij}^{2}}}) \exp(-\frac{(\alpha_{ij}-\frac{q_{i}k_{j}}{\sqrt{d}})^{2}}{2 \sigma_{ij}^{2}}).
\label{eq:loss_UPD}
\end{equation} 

As a summary, our UPD module captures and models the uncertainty of complex driving environments by representing attention scores as  Gaussian distributions, which are more robust and capable of handling variations and noise in the 3D MOT, 
and the \cref{eq:det} can be improved as: 
\begin{equation}
   \textbf{B}_t,\textbf{Q}_t\leftarrow \mathbf{Decoder}_\text{\UPD{}}(\textbf{F}_t,\textbf{Q}_t).
   \label{eq:det_our}
\end{equation}

\input{latex/table/main_res_val}

\subsection{\denoise{}}
\label{subsec:denoise}
In complex 3D MOT scenarios, challenges such as occlusions and varying object sizes can hinder the learning and convergence of query-based methods. The slow convergence and suboptimal results from the instability of bipartite graph matching.
To address this challenge, we draw inspiration from DN-DETR~\cite{li2022dn} and propose \textbf{H}ierarchical \textbf{Q}uery \textbf{D}enoising (\HQD{}) training strategy, which incorporates query denoising to enhance training process for stable optimization.
We perturb GT bounding boxes with noises into the decoder and train the model to reconstruct the original boxes, which effectively reduces graph matching difficulty and leads to faster convergence.

Specifically, we start by perturbing the ground truth boxes to generate noised queries. To enhance the model's ability to handle various complex driving scenarios, we define hierarchical challenging levels for the perturbed queries.
We set lower and upper bound thresholds, denoted as $\beta_{\text{lower}}$ and $\beta_{\text{upper}}$ respectively, to help categorize the noised queries into three classes based on their challenging levels.
We identified positive samples ({\fontsize{9}{10}\selectfont "Pos"} in \cref{fig:main_pipline}), i.e., low-challenging samples, when the 3D Intersection over Union (IoU) between a noised query and its corresponding ground truth exceeds the $\beta_{\text{upper}}$ threshold, i.e., $IoU_{\textbf{q}_t^i}>\beta_{\text{upper}}$. Conversely, negative samples ({\fontsize{9}{10}\selectfont "Neg"} in \cref{fig:main_pipline}), i.e., high-challenging samples, are defined when the 3D IoU falls below the $\beta_{\text{lower}}$ threshold, i.e., $IoU_{\textbf{q}_t^i}<\beta_{\text{lower}}$. 
Intermediate IoU values are disregarded ({\fontsize{9}{10}\selectfont "Ign"} in \cref{fig:main_pipline}), as they do not provide clear indications of any challenging factors, and can disrupt the normal query learning process as demonstrated in \cref{tab: uncertainty similarity learning upper bound threshold}.
The resulting set of noised queries that meet these requirements is denoted as $\textbf{Q}_\text{HQD-noised}$, and then the decoder process 
can be expanded as:
\begin{equation}
   \textbf{B}_t^{\HQD{}},\textbf{Q}_t^{\HQD{}}\leftarrow \mathbf{Decoder}(\textbf{F}_t, \textbf{Q}_\text{\HQD{}-noised}).
   \label{eq:det_our2}
\end{equation}
For optimization, the loss for positive samples and negative samples are calculated to form the target:
\begin{equation}
\mathcal{L}_{\HQD{}} =  \mathcal{L}_{box}^{pos} + \mathcal{L}_{cls}^{pos} + \mathcal{L}_{cls}^{neg},
\label{eq:loss_HQD}
\end{equation}
where $\mathcal{L}_{cls}^{pos}$ and $\mathcal{L}_{box}^{pos}$ are respectively focal loss~\cite{lin2017focal} and L1 loss for the classification and box loss of $\textbf{B}_t^{U}$, while $\mathcal{L}_{cls}^{neg}$ is focal loss to distinguish the background.


By incorporating the query denoising and handling the noised queries based on their noise/challenging levels, our HQD module enhances the training robustness and convergence, leading to more stable and accurate results.

\subsection{Overall Optimization}
To summarize, the overall optimization target of our \name{} is formulated as:
\begin{equation}
\begin{aligned}
\mathcal{L} = & \lambda_{tracking} \mathcal{L}_{tracking}  + \lambda_{\PQI{}}  \mathcal{L}_{\PQI{}}  \\ & +  \lambda_{\UPD{}}  \mathcal{L}_{\UPD{}} + \lambda_{\HQD{}}  \mathcal{L}_{\HQD{}},
\end{aligned}
\end{equation}
where $\mathcal{L}_{tracking} = \mathcal{L}_{cls} +  \mathcal{L}_{box}$, $\mathcal{L}_{cls}$ and $\mathcal{L}_{box}$ are classification loss and box loss for the tracked objects, $\mathcal{L}_{\PQI{}}$, $ \mathcal{L}_{\UPD{}}$, and $\mathcal{L}_{\HQD{}}$  are defined in \cref{eq:losscei}, \cref{eq:loss_UPD}, and \cref{eq:loss_HQD}, and $\lambda$ indicate the weight balance coefficients that are all set to 1.0 by default.

%% file: latex/fig/attention.tex
\begin{figure}[ht]
  \centering
  \includegraphics[width=1\linewidth]{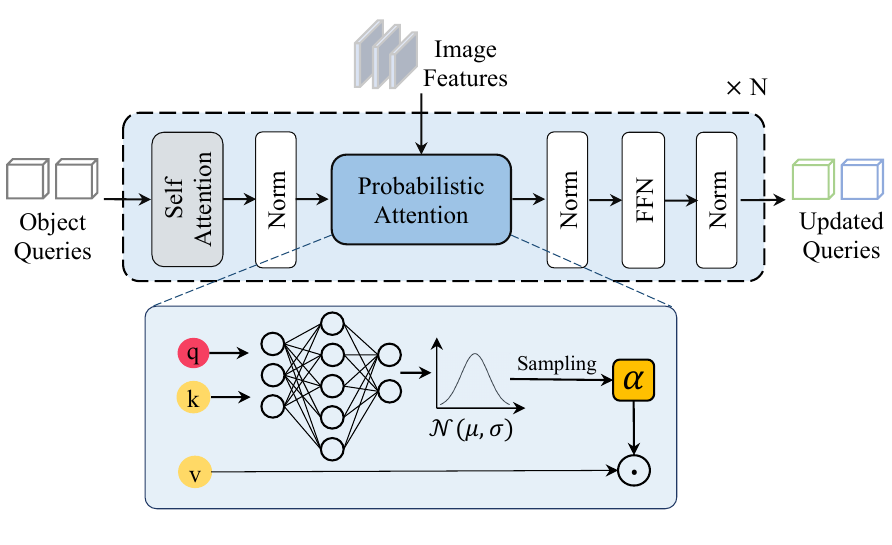}
  \vspace{-12pt}
  \caption{\textbf{\decoder{} (\UPD{}) architecture}. 
  The traditional cross-attention is upgraded with probabilistic attention to quantifying the uncertainty. The probabilistic attention utilizes a multi-layer perception that takes the query $q$ and key $k$ as input to generate the mean and standard deviation, which are used to form a Gaussian distribution. Subsequently, the attention value $\alpha$ is sampled from the constructed Gaussian distribution. 
  }
  \vspace{-12pt}
  \label{fig:attention}
\end{figure}

%% file: latex/table/main_res_val.tex
\begin{table*}[tb]

   \caption{
  \textbf{Comparisons with previous methods on the nuScenes \textit{val} set}. Our \name{} outperforms all existing camera-based 3D MOT methods in all metrics.
  }
   \vspace{-5pt}
  \centering
  \addtolength{\tabcolsep}{-2.2pt}
 \begin{tabularx}{1\textwidth}{l|c|cc|cccc}
    \toprule
  Method &  Backbone &  AMOTA$\uparrow$ &  AMOTP$\downarrow$ &  RECALL$\uparrow$ &  MOTA$\uparrow$ &  MOTP $\downarrow$ &  IDS$\downarrow$ \\
    \midrule
      \midrule
    DEFTT~\pub{arxiv21}~\cite{chaabane2021deft} & DLA-34 & 0.201 & -- & -- & 0.171 &-- & -- \\
    QD-3DT~\pub{TPAMI2022}~\cite{hu2022QD3DT} & DLA-34 &0.242 & 1.518 & 39.9\% & 0.218 & --&5646 \\
    MSGLMB\pub{ICCAIS24}~\cite{van2024msglmb} & -- & 0.382 & 1.235 & 55.6\% & 0.399 & -- & 2929 \\
    CC-3DT~\pub{CoRL2022}~\cite{cc3dt}& R101 & 0.429 & 1.257 & 53.4\% & 0.385& --& 2219 \\
    Cyclic~\pub{IJCV24}~\cite{guo2024cyclic} & R101 & 0.469 & 1.002 & 45.7\% & 0.354 & -- & 3613\\
    QTrack~\pub{arxiv23}~\cite{yang2022QTtrack} & V2-99 & 0.511 & 1.090 & 58.5\%& 0.465 &--& 1144\\
    RockTrack~\pub{arxiv24}~\cite{li2024rocktrack} & -- & 0.514 & 1.144 & -- &	--	& -- &	--	\\

    \midrule
    \multicolumn{3}{l}{\textit{\textbf{Tracking with Query}}} \\
    \midrule    MUTR3D~\pub{CVPR2022}~\cite{zhang2022mutr3d} & R101 & 0.294 & 1.498 & 42.7\% & 0.267& 0.709 & 3822 \\
    STAR-TRACK~\pub{RA-L2023}~\cite{doll2023star} & R101 & 0.379 & 1.358 & 50.1\% & 0.360 &--& 372\\
     Sparse4D-v3~\pub{arxiv23}~\cite{lin2023sparse4d} & R101 & 0.567 & 1.027 & 65.8\% & 0.515 &0.621& 557 \\
    DQTrack~\pub{ICCV23}~\cite{li2023dqtrack} &V2-99 & 0.446 & 1.251 & 62.2\% & -- & -- & 1193 \\
    PF-Track-S~\pub{CVPR2023}~\cite{pang2023PFtrack} & V2-99 & 0.408 & 1.343 & 50.7\% & 0.376 &--& 166 \\
    HSTrack~\pub{arxiv24}~\cite{lin2024hstrack} & V2-99 & 0.464 & 1.262 & 56.9\% & 0.423 &--& 204 \\
    PF-Track-F~\pub{CVPR2023}~\cite{pang2023PFtrack}& V2-99 & 0.479 & 1.227 & 59.0\% & 0.435 &--& 181 \\
    ADA-Track~\pub{CVPR2024}~\cite{ding2024adatrack}& V2-99 & 0.479 & 1.246 & 60.2\% & 0.430 &--& 767 \\
   \rowcolor{mygray} \textbf{\name-F} (Ours) & V2-99& 0.566& 1.090 &62.2\%&0.498  &0.657 & 174\\
   \rowcolor{mygray} \textbf{\name-S} (Ours)& ViT-L  & 0.600 & 1.020 & 68.8\% & 0.538 & 0.614 & 167 \\
   \rowcolor{mygray} \textbf{\name-F} (Ours) & ViT-L    & \bf0.652 & \bf0.924 & \bf72.2\% & \bf0.574 & \bf0.577& \bf134\\
    \bottomrule
      \multicolumn{8}{l}{\footnotesize{"\textbf{S}" and "\textbf{F}" represent the settings of small-resolution and full-resolution respectively.} }
  \end{tabularx}

  \label{tab:main_results_validate}
  \vspace{-10pt}
\end{table*}

%% file: sec/4_exp.tex
\section{Experiment}

\subsection{Experimental Setup}
\label{subsec: setup}
We conduct experiments on the large-scale nuScenes benchmark~\cite{caesar2020nuscenes} and follow the official evaluation metrics from nuScenes.
Detailed Metric and Implementation are present in the Appendix.

\subsection{State-of-the-art Comparison}
\label{subsec: main_res}

\input{latex/table/main_res_test}
\mypara{Tracking on nuScenes val set.}
In \cref{tab:main_results_validate}, we compare our \name{} with state-of-the-art methods on nuScenes val set. 
First, our method significantly outperforms existing algorithms across all tracking metrics, whether they are end-to-end or non-end-to-end methods. Specifically, \name{} achieves impressive performance with 65.2\% AMOTA and 0.924 AMOTP. 
When compared with the previous query-based tracker Sparse4D-v3, the performance gap is further enlarged to 8.5\% AMOTA.
Second, we validate the generality of \name{} by applying different encoder backbones, i.e., V2-99 and ViT. Equipped with V2-99, the proposed framework achieves consistent gains with 8.7\% AMOTA (\name-F 0.566 vs PF-Track-F 0.479 vs ADA-Track 0.479).
Moreover, when employing the larger ViT-L as our backbone, we further achieve leading performance.

\mypara{Tracking on nuScenes test set.}
In \cref{tab:main_results_test}, we compare our \name{} with state-of-the-art camera-based methods on nuScenes test set. Our proposed \name{} maintains an end-to-end tracking pipeline without heuristic post-processing and achieves leading performance with 66.3\% AMOTA, surpassing the previous best solution Sparse4D-v3 by a significant margin of 8.9\% AMOTA.

\subsection{Analysis Study}
\label{subsec: uncertainty}

\mypara{Analysis and ablations of the proposed modules of \name{}.}
\input{latex/table/ablation}
In \cref{tab: modules ablation}, we validate our proposed modules for our model’s performance on nuScenes val set.
It is clear that incorporating each module leads to performance gain in tracking. Specifically, the Uncertainty-aware Probabilistic
Decoder (\UPD{}) module significantly improves the baseline with 4.4\% AMOTA, and the 2D-Prompted Query Initialization (\PQI{}) and Hierarchical Query Denoising (\HQD{}) modules obtained consist boost with 2.4\% and 2.9\% AMOTA respectively.
Moreover, combining the three modules leads to further improvements.

\mypara{Analysis of uncertainty.}
Quantifying uncertainty can be challenging for transformer-based models which do not inherently provide uncertainty estimates like Bayesian methods.
Thus, we can only employ computationally intensive strategies to approximate uncertainty, such as the Monte Carlo Dropout (MC dropout)~\cite{gal2016dropout} and Ensemble strategy (Ens.)~\cite{lakshminarayanan2017simple}. 
As shown in \cref{tab: modules ablation}, the proposed UPD module successfully reduces uncertainty. Surprisingly, other modules also effectively reduce uncertainty, even though they were not designed to aim at uncertainty.

\input{latex/table/uncertainty}

\input{latex/table/uncertainty_ths}
\input{latex/table/depthnet}
\input{latex/table/different_arch}

\mypara{Analysis of complex situations.}
In \cref{tab:uncertainty}, we analyze the performance of \name{} under different driving conditions, i.e., different visibilities, different object sizes, and different distances.
Our \name{} achieves consistent improvements over PF-Track~\cite{pang2023PFtrack} under all visibility, object size, and distance settings. Furthermore, \name{} brings larger improvements in more challenging situations, such as lower visibilities (+ 6.3\% AMOTA), smaller objects (+ 5.5\% AMOTA), and more distant objects (+ 4.1\% AMOTA), both of which demonstrate our effectiveness in addressing diverse challenges in 3D MOT.

\subsection{Ablation Study}
\label{subsec: ablations}

\subsubsection{Ablations on thresholds of HQD module.}

In \cref{tab: uncertainty similarity learning upper bound threshold}, we validate various settings of the threshold of HQD module.  (1) In the first setting, we do not consider the noise thresholds, meaning that all noised queries undergo box optimization. (2) In the second setting, we do not selectively denoise queries based on hierarchical challenging levels; instead, a single threshold is used to classify all noised queries into positive and negative samples for optimization. (3) In our HQD module, we selectively denoise queries based on hierarchical challenging levels. 
The improved performance demonstrates our HQD, which optimizes samples based on different challenging levels, effectively handles various complex driving environments, and achieves superior results.
We also investigate varying lower and upper bound thresholds, i.e., $\beta_{\text{lower}}$ and $\beta_{\text{upper}}$.
The model achieves the best result when $\beta_{\text{lower}}$ is set to 0.30 and $\beta_{\text{upper}}$ is set to 0.70.

\subsubsection{Ablations on network strides of PQI module.} 

In \cref{tab: stride of DepthNet}, we conduct an ablation study to analyze the effects of the network stride of PQI module. As described in \cref{eq:depth}, our PQI module incorporates two additional auxiliary tasks based on the extracted feature $\textbf{F}_t$. To reduce computational overhead, we directly reuse features from different layers with varying strides in the feature pyramid network, thereby alleviating the burden on the auxiliary task heads. We present the performance achieved with different strides for the image features. The experimental results reveal that the model performs optimally when using a stride of 16.

\subsubsection{Ablations on different decoders.}
In \cref{tab: different detection methods}, we validate the generality of \name{} by applying different decoders. 
Specifically, we employ two popular decoders: PETR~\cite{liu2022petr} and DETR3D~\cite{wang2022detr3d}, which are widely adopted by query-based tracking methods. Our approach achieves excellent results with both decoders, yielding comparable performance. The results obtained with DETR3D are slightly higher, leading us to select DETR3D as our default decoder.

\subsection{Qualitative Comparison}
\label{subsec: vis}
In \cref{fig: qualitative results}, we provide the qualitative results on nuScenes dataset. We compare our \name{} with the previous state-of-the-art end-to-end tracker, PF-Track~\cite{pang2023PFtrack}.
In \cref{fig: qualitative results} (a), PF-Track exhibits commendable tracking accuracy when the line of sight is clear at time $t_{i}$. However, as the occlusion gradually intensifies, the accumulated error arising significantly escalates. Especially at $t_{i+12}$, the predicted bounding boxes for two pedestrians completely overlap. In contrast, our \name{} consistently maintains a high level of tracking precision throughout the entire duration of continuous tracking.

In complex scenarios of \cref{fig: qualitative results} (b), which are characterized by multiple complex factors such as occlusions in crowded and spacious environments, as well as the small size of vehicles and pedestrians, \name{} achieves more precise tracking bounding boxes and successfully recognizes more tracked objects compared to PF-Track~\cite{pang2023PFtrack}.
Furthermore, the visualization of our attention scores demonstrates a higher concentration on the object center, indicating that our model pays more attention to the target objects. In contrast, PF-Track~\cite{pang2023PFtrack} exhibits low attention scores for challenging samples, failing to capture and track objects.
The outstanding 3D MOT results of \name{} demonstrate the effectiveness of our framework, which addresses the challenges across various complex tracking scenarios.

\input{latex/fig/results}

%% file: latex/table/main_res_test.tex
\begin{table*}[tb]

     \caption{
  \textbf{Comparisons with state-of-the-art camera-based methods on the nuScenes \textit{test} set}. 
  }
   \vspace{-5pt}
  \centering
   \addtolength{\tabcolsep}{-2.9pt}
 \begin{tabularx}{1\textwidth}{l|c|cc|cccc}
    \toprule
    Method & Backbone & AMOTA$\uparrow$ & AMOTP$\downarrow$ & RECALL$\uparrow$ & MOTA$\uparrow$ & MOTP $\downarrow$ & IDS$\downarrow$ \\
    \midrule
    \midrule

     PermaTrack~\pub{ICCV2021}~\cite{tokmakov2021learning} & DLA-34 & 0.066 & 1.491 & 18.9\% & 0.060 & 0.724& 3598 \\
     
     DEFT~\pub{arxiv21}~\cite{chaabane2021deft} &  DLA-34 &  0.177 & 1.564 & 33.8\% & 0.156 &0.770 & 6901 \\
     QD-3DT~\pub{TPAMI2022}~\cite{hu2022QD3DT} & DLA-34 & 0.217 & 1.550 & 37.5\% & 0.198 &0.773 & 6856 \\
     CC-3DT~\pub{CoRL2022}~\cite{cc3dt} &  R101 & 0.410 & 1.274 & 53.8\% & 0.357 & 0.676& 3334 \\
     Cyclic~\pub{IJCV24}~\cite{guo2024cyclic} & R101 & 0.433& 1.055 & 49.2\%& 0.334 & -- &6621 \\
    QTrack~\pub{arxiv23}~\cite{yang2022QTtrack} & V2-99 & 0.480 & 1.100 & 58.3\% & 0.431 & 0.597& 1484 \\

        \midrule
    \multicolumn{3}{l}{\textit{\textbf{Tracking with Query}}} \\
    \midrule

    MUTR3D~\pub{CVPR2022}~\cite{zhang2022mutr3d} &  R101 & 0.270 & 1.494 & 41.1\% & 0.245 & 0.709 & 6018 \\
    
    PF-Track-F~\pub{CVPR2023}~\cite{pang2023PFtrack}  &  V2-99 & 0.434 & 1.252 & 53.8\% & 0.378 &0.674 & \bf 249 \\
    STAR-TRACK~\pub{RA-L2023}~\cite{doll2023star} & V2-99 & 0.439 & 1.256 & 56.2\% & 0.406 &0.664 & 607\\
    
    DQTack~\pub{ICCV23}~\cite{li2023dqtrack} & V2-99 &0.523 & 1.096 & 62.2\% & 0.444 & 0.649 & 1204 \\

    Sparse4D-v3~\pub{arxiv23}~\cite{lin2023sparse4d} & V2-99 & 0.574 & 0.970 & 66.9\% & 0.521 & \textbf{0.525} & 669 \\
    ADA-Track~\pub{CVPR2024}~\cite{ding2024adatrack}& V2-99 & 0.456 & 1.237 & 55.9\% & 0.406 &--& 834 \\
 \rowcolor{mygray}  \textbf{\name-F} (Ours) & V2-99 & 0.608 & 0.925 & \bf75.8\% & 0.547 & 0.559 & 963  \\
     \rowcolor{mygray}  \textbf{\name-F} (Ours) & ViT-L & \bf 0.663 & \bf 0.815 &  72.3\% &\bf  0.554 & 0.530 & 844 \\

    \bottomrule
    \multicolumn{8}{l}{\footnotesize{"\textbf{F}" represent on the full-resolution settings.} }
   \end{tabularx}

    \label{tab:main_results_test}
  \vspace{-5pt}
\end{table*}

%% file: latex/table/ablation.tex
\begin{table*}[tb]
   
    \caption{
  \textbf{Analysis and ablations on the proposed modules of \name{}.} 
  }
 \vspace{-5pt}
  \centering
   \addtolength{\tabcolsep}{0.8pt}
 \begin{tabularx}{1\textwidth}{ccc|cccccc|cccc}
    \toprule
    \multicolumn{3}{c|}{Method} & \multicolumn{6}{c|}{Tracking} & \multicolumn{3}{c}{Uncertainty}
    \\
    \midrule
    
    \UPD{} &  \PQI{} & \HQD{} & AMOTA$\uparrow$ & AMOTP$\downarrow$ & RECALL$\uparrow $& MOTA$\uparrow$ & MOTP$\downarrow$ & IDS$\downarrow$ & & \cellcolor{mypink} $s$$\downarrow$&\cellcolor{mypink} $\sigma$$\downarrow$\\
            
     \midrule

     \xmarkg & \xmarkg & \xmarkg &  0.394 & 1.363 & 51.4\% & 0.372& 0.753 & 178 & \multirow{6}{*}{\rotatebox{90}{{MC Dropout}}}  & 1.99 & 0.108 \\ 
     \colorbox{mycyan}{\cmark} &  \xmarkg & \xmarkg  &  0.438& 1.261 & 55.9\% & 0.419& 0.681 & 175  &  &1.86 & 0.085  \\
     \xmarkg &  \colorbox{mycyan}{\cmark} & \xmarkg  &  0.418 & 1.251 & 54.6\% & 0.394 & 0.667 & 177 &  & 1.91 & 0.097 \\
      \xmarkg &  \xmarkg & \colorbox{mycyan}{\cmark}  &  0.423 & 1.264 & 55.6\% & 0.397 & 0.671 & 183  & &  1.88 & 0.093 \\

        \colorbox{mycyan}{\cmark} & \colorbox{mycyan}{\cmark} & \colorbox{mycyan}{\cmark} &   \cellcolor{mygray} \textbf{0.458}&   \cellcolor{mygray} \textbf{1.230} &   \cellcolor{mygray}\textbf{56.6\%} &   \cellcolor{mygray}\textbf{0.433} &    \cellcolor{mygray}\bf 0.664 &   \cellcolor{mygray}\textbf{172} & &   \cellcolor{mygray} \bf 1.81 &   \cellcolor{mygray} \bf 0.078\\

      \midrule

  \multicolumn{3}{c|}{PF-Track-S}  & 0.408 & 1.343 & 50.7\% & 0.376 &--& 166& \multirow{2}{*}{\rotatebox{90}{{Ens.}}}  & 1.96 & 0.100  \\
 \multicolumn{3}{c|}{\cellcolor{mygray}\textbf{\name-S}}&    \cellcolor{mygray} \bf 0.458&    \cellcolor{mygray} \bf 1.230 &   \cellcolor{mygray} \bf 56.6\%&    \cellcolor{mygray} \textbf{0.433} &    \cellcolor{mygray} \bf 0.664 &   \cellcolor{mygray} \textbf{172} & &   \cellcolor{mygray} \bf 1.75 &    \cellcolor{mygray}\bf 0.072 \\

    \bottomrule
\multicolumn{8}{l}{\footnotesize{$s$ and $\sigma$ donate entropy and standard deviation of uncertainty quantification, respectively.} }
    
  \end{tabularx}

  \label{tab: modules ablation}
\vspace{-8pt}

\end{table*}

%% file: latex/table/uncertainty.tex
\begin{table}[ht]

    \caption{
  \textbf{Analysis of complex situations.}
  }
    \vspace{-5pt}
  \centering
  \addtolength{\tabcolsep}{2.15pt}
 \begin{tabularx}{1\linewidth}{l|ccc}
    \toprule
  \bf Visibilities  &0-40\%&40\%-60\%&60\%-100\% \\
    \midrule
    PF-Track & 38.3 & 38.6& 39.5 \\
\rowcolor{mygray}\bf  \name{} &  44.6 \textcolor{purple}{\scalebox{0.8}[0.8]{(\textbf{+6.3})}} & 44.8 \scalebox{0.8}[0.8]{(\textbf{+6.2})} & 45.2 \scalebox{0.8}[0.8]{(\textbf{+5.7})} \\
    \midrule
\bf Size & 0-2m &  2-3.5m & \textgreater3.5m\\
    \midrule
PF-Track& 22.8 & 20.4 & 19.1    \\
\rowcolor{mygray} \bf \name{}& 28.3 \textcolor{purple}{\scalebox{0.8}[0.8]{(\textbf{+5.5})}} & 25.5 \scalebox{0.8}[0.8]{(\textbf{+5.1})} & 24.0 \scalebox{0.8}[0.8]{(\textbf{+4.9})}\\
    \midrule
\bf Distance & \textgreater30m & 20-30m  &0-20m \\
     \midrule
 PF-Track&7.5& 38.9  &64.6 \\
\rowcolor{mygray} \bf \name{} & 11.6 \textcolor{purple}{\scalebox{0.8}[0.8]{(\textbf{+4.1})}} & 42.8 \scalebox{0.8}[0.8]{(\textbf{+3.9})} &67.0 \scalebox{0.8}[0.8]{(\textbf{+2.4})}\\

    \bottomrule
          \multicolumn{4}{l}{\footnotesize{The metric is AMOTA$\uparrow$.} }
  \end{tabularx}

  \vspace{-7pt}
    \label{tab:uncertainty}
\end{table}

%% file: latex/table/uncertainty_ths.tex
\begin{table}[ht]

    \caption{
  \textbf{Ablations on thresholds of HQD}. (1) Without thresholds, (2) without selecting by levels, (3) Our HQD.
  }
  \vspace{-5pt}
  \centering
   \addtolength{\tabcolsep}{-4.7pt}
 \begin{tabularx}{1\linewidth}{c|ccccc}
    \toprule
    Threshold & AMOTA$\uparrow$ & AMOTP$\downarrow$ & RECALL$\uparrow$ & MOTA$\uparrow$  & IDS$\downarrow$ \\
    \midrule
    (1) & 0.440 & 1.249 & 55.9\% & 0.419  & 181 \\     
     (2) & 0.445 & 1.239 & 56.5\% & 0.425  &174 \\     
      \rowcolor{mygray}   \underline{(3)} & \bf0.458 & \bf1.230 & \bf56.6\% & \bf0.433  & \bf172 \\
     \midrule
    \multicolumn{3}{l}{\textit{{Lower Bound Thresholds}}} \\
     \midrule
    0.20 & 0.450 & 1.236 & 56.3\% & 0.429  & 179 \\
\rowcolor{mygray}     
   \underline{0.30} & \bf0.458 & \bf1.230 & \bf56.6\% & \bf0.433 & \bf172 \\
    0.40 & 0.448 & 1.241 & 56.4\% & 0.427 & 173 \\
     \midrule
    \multicolumn{3}{l}{\textit{{Upper Bound Thresholds}}}\\
     \midrule
 0.60 & 0.436 & 1.261 & 56.1\% & 0.419   &204 \\
    0.65 & 0.455 & 1.243 & \bf57.5\% & \bf0.434  &198 \\
  \rowcolor{mygray}   \underline{0.70} & \bf0.458 & \bf1.230 & 56.6\% & 0.433  &\bf172 \\
    0.75 & 0.444 & 1.253 & 55.9\% & 0.416  & 214 \\
    \bottomrule
  \end{tabularx}

    \label{tab: uncertainty similarity learning upper bound threshold}
  \vspace{-7pt}
\end{table}

%% file: latex/table/depthnet.tex

\begin{table}[th]

    \caption{
  \textbf{Ablations on network strides of PQI module}. 
  }
    \vspace{-5pt}
  \centering
  \addtolength{\tabcolsep}{-3.4pt}
 \begin{tabularx}{1\linewidth}{c|cccccc}
    \toprule
    Stride & AMOTA$\uparrow$ & AMOTP$\downarrow$ & RECALL$\uparrow$ & MOTA$\uparrow$   & IDS$\downarrow$ \\
    \midrule
    8 & 0.456 & 1.245 & 56.1\% & 0.422   &178 \\
\rowcolor{mygray}     \underline{16} & \bf0.458 & \bf1.230 & \bf56.6\% & \bf0.433   & \bf172 \\
    32 & 0.446 & 1.268 & 56.2\% & 0.419 &214 \\
    \bottomrule
  \end{tabularx}
  
  \label{tab: stride of DepthNet}
  \vspace{-7pt}
\end{table}

%% file: latex/table/different_arch.tex
\begin{table}[h!]
    
    \caption{
  \textbf{Ablations on different decoders}.
  }
\vspace{-5pt}
  \centering
  \addtolength{\tabcolsep}{-4.5pt}
 \begin{tabularx}{1\linewidth}{l|ccccc}
    \toprule
     Decoder & AMOTA$\uparrow$ & AMOTP$\downarrow$ & RECALL$\uparrow$ & MOTA$\uparrow$  & IDS$\downarrow$ \\
    \midrule
     PETR & 0.452 & 1.246 & \bf57.1\% & 0.429  & 196 \\
\rowcolor{mygray}      \underline{DETR3D} & \bf0.458 & \bf1.230 & 56.6\% & \bf0.433  & \bf172 \\
    \bottomrule
  \end{tabularx}

  \label{tab: different detection methods}
  \vspace{-7pt}
\end{table}

%% file: latex/fig/results.tex
\begin{figure}[h!]
  \centering
  \includegraphics[width=\linewidth]{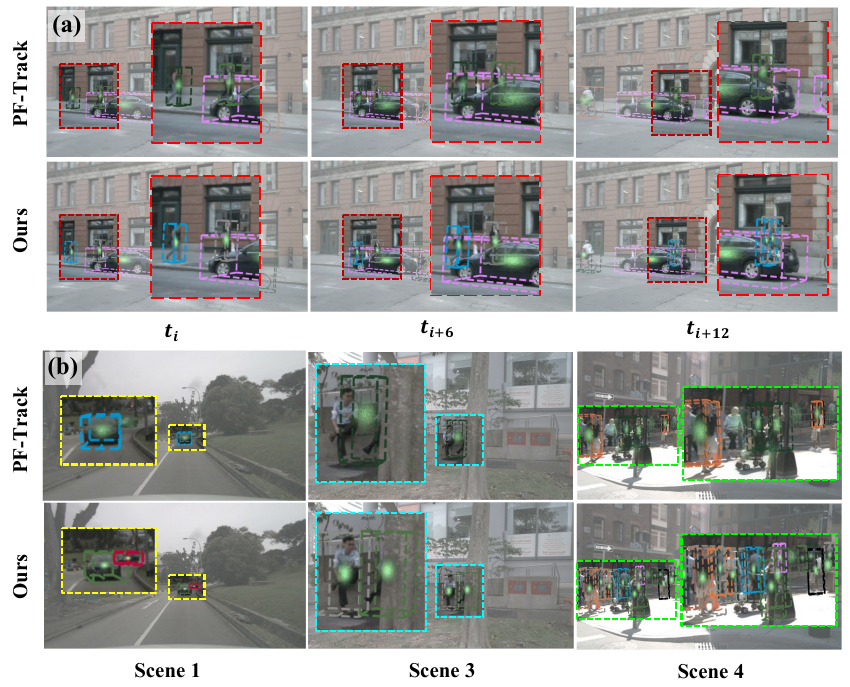}
\caption{\textbf{Qualitative results on the nuScenes dataset}.
  \textbf{(a)} The tracking results for an occlusion scenario of two pedestrians of consecutive frames ($t_i-t_{i+12}$). 
  \textbf{(b)} The tracking results on several challenging tracking scenes. Moreover, we plot the attention scores of object queries, which indicate how strongly the model focuses on the target objects. A higher concentration of color represents a higher attention score and a stronger confidence in the corresponding object. 
  \vspace{-7pt}
  }
  \label{fig: qualitative results}
  \vspace{-12pt}
\end{figure}

%% file: sec/5_conclusion.tex
\section{Conclusion}
In this paper, we improve the current end-to-end 3D MOT framework from multiple aspects. 
Specifically, for query initialization, we propose \init{} to improve object localization accuracy by incorporating predicted 2D location and depth cues.
For query propagation, the \decoder{} models the uncertainty of complex driving situations using probabilistic attention, providing a comprehensive understanding of predictive uncertainty. 
Then, for query matching, the \denoise{} strategy enhances training robustness and convergence.  
Experimental results on nuScenes benchmark demonstrate that \name{} achieves state-of-the-art performance, i.e., 66.3\% AMOTA on the test split, with a significant improvement of 8.9\% AMOTA.

%% file: sec/supp.tex
\clearpage
\newpage



\begin{center}
\noindent{\textbf{\large{Appendix}}}
\end{center}



\section{Additional Details}
\subsection{Depthnet Details}
\begin{figure}[h!]
\centering
\includegraphics[width=0.32\textwidth]{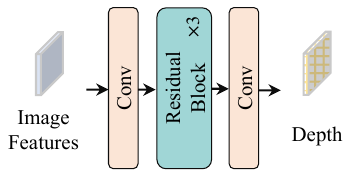}
\caption{\textbf{Details of the depth network in \PQI{}.}}
\label{fig: depthnet}
\end{figure}
The depth network of the proposed PQI module is composed of multiple residual blocks, and we provide an illustration in \cref{fig: depthnet}.

\subsection{Experimental Setup}
\label{subsec: setup2}
\subsubsection{Dataset.}
We conduct experiments on the popular nuScenes benchmark \cite{caesar2020nuscenes}, which is a large-scale autonomous-driving dataset for 3D detection and tracking, consisting of 700, 150, and 150 scenes for training, validation, and testing, respectively. Each frame contains one point cloud and six calibrated images from the surrounding cameras with a full 360-degree field of view.
It provides 3D tracking bounding boxes from 7 categories for the tracking task.

\subsubsection{Metrics.}
We follow the official evaluation metrics from nuScenes.
For the 3D tracking task, we report Average Multi-object Tracking Accuracy (AMOTA)~\cite{weng2019baseline}, Average Multi-object Tracking Precision (AMOTP), and the modified CLEAR MOT metrics~\cite{bernardin2008evaluating}, e.g., MOTA, MOTP, and IDS. For a detailed understanding, please refer to \cite{caesar2020nuscenes, bernardin2008evaluating}.

\subsubsection{Implementation details.}
In this paper, we assess the generalization capability of our \name{} through experiments using different encoders, e.g., V2-99~\cite{lee2019energy} and ViT~\cite{dosovitskiy2020image}, and different decoders e.g., PETR~\cite{liu2022petr} and DETR3D~\cite{wang2022detr3d}. 
All experiments are conducted on 8 NVIDIA A100-80GB GPUs. 
For each training sample, it contains three consecutive adjacent frames each with contains six surrounding images, and we use a fixed number of 500 initial queries for each sample.
We adopt the AdamW optimizer~\cite{loshchilov2017decoupled} for network training, with the initial learning rate setting of 0.01 and the cosine weight decay set to 0.001. By default, the thresholds $\beta_{\text{lower}}$ and $\beta_{\text{upper}}$ are set to 0.3 and 0.7, and the weight coefficients $\lambda$ that are all set to 1.0, respectively.

Due to the limited computation resources, we follow PF-Track~\cite{pang2023PFtrack} to apply two resolution settings, full-resolution and small-resolution. For full-resolution (“-F”), we crop the origin $1600 \times 900$ image to $1600 \times 640$. For small-resolution (“-S”), we scale down the cropped image to 800 x 320 in a further step.
We pre-train the image backbone with single-frame detection task for 12 epochs (small-resolution setting) and 24 epochs (full-resolution setting) respectively, and further train the end-to-end tracker with consecutive frames (set to be 3 frames) for another 12 epochs (small-resolution) and 24 epochs (full-resolution). All the ablation studies are conducted on the small-resolution setting with V2-99 backbone.

We ulitze the 3D location to initialize queries by following steps: 1) normalize input coordinates to the [0, 2$\pi$] range; 2) generate frequency bands using exponential temperature scaling; 3) compute sine/cosine components for each dimension (X, Y, Z); 4) concatenate the encoded dimensions; 5) project the concatenated features through two linear layers with ReLU activation. We will include these implementations in the revision.

\subsubsection{Analysis study details.}
The analysis studies in Tab. 4 are conducted on the small-resolution setting with V2-99 backbone.
The experiments are performed on the nuScenes val set, in which we focus on specific challenging conditions to select clips for evaluation.
We utilize the attributes of the bounding boxes provided by the nuScenes dataset, e.g., the visibility labels, and then calculate the average for each clip, finally group the results according to different ranges of the attributes. 
The categorization process involved the following criteria:
\noindent\textit{1. Different visibilities}: 
the dataset is divided based on the visibility attribute of the objects. Visibility ranges are considered as 0-40\%, 40-60\%, and 60-100\%. 
\noindent\textit{2. Different object sizes}: 
the dataset is divided into three groups based on the average object size: objects with a size greater than or equal to 3.5 meters, objects smaller than 3.5 meters and greater than or equal to 2 meters, and objects smaller than 2 meters.
\noindent\textit{3. Different object distances}: 
the dataset is split based on different distance ranges, namely 0-20 meters, 20-30 meters, and 30 meters and above.
By applying these categorizing and calculations, subsets of data were selected from the clips in the validation dataset to evaluate their performance based on the specified challenging conditions.

\section{Additional Discussion}
\subsection{PQI vs. 3DPPE}
Previous 3DPPE~\cite{shu20233dppe} also involves depth priors in a query-based framework, it differs from S2-Track in several aspects. First, 3DPPE focuses on 3D object detection, whereas we tackle 3D MOT. Second, 3DPPE introduces 3D point positional encoding, while our PQI is designed for query initialization. Moreover, we also retain randomly initialized queries to explore missing objects. We will add this discussion into the revision.

\subsection{Analysis of other modules also effectively reduce the uncertainty}
Our PQI module leverages learned certain priors, i.e., 2D object location and depth information, to enhance the initialization of queries, thus effectively reducing the uncertainty in query initialization and resulting in more accurate object localization and tracking.
The HQD strategy introduces different levels of noise to the queries and then applies a denoising process, allowing the model to encounter varying magnitudes of noise (i.e., uncertainty) during training. This effectively helps the model reduce uncertainty during query matching, leading to more stable and accurate tracking performance. Although the motivation of these two modules is not uncertainty, they both help the model reduce uncertainty during query initialization and matching. Moreover, they are incorporated together with the UPD module, which aims to reduce uncertainty during query propagation.

\section{Additional Results}
\subsection{Inference Latency}


We present the inference latency measure in \cref{tab:latency}. Our proposed \name{}, demonstrates greater efficiency compared to previous end-to-end methods, i.e., MUTR3D~\cite{zhang2022mutr3d} and DQTrack~\cite{li2023dqtrack}. 
In comparison to the state-of-the-art PF-Track~\cite{pang2023PFtrack}, our \name{} introduces little additional latency, and yet this trade-off results in a 5.0\% improvement in AMOTA over PF-Track. Further efficiency enhancement in tracking is a promising direction for future research.

\begin{table}[ht]
\caption{\textbf{Inference latency}. Frame Per Second (FPS) is evaluated on a single NVIDIA A100 GPU from input images to tracking results.}
	\centering
       \setlength\tabcolsep{18.5pt}
       \begin{tabularx}{0.45\textwidth}{c|c}
           \toprule
           Method & FPS \\
           \toprule
           PF-Track-S~\cite{pang2023PFtrack}  &  9.2 \\
           DQTrack~\cite{li2023dqtrack} & 6.0 \\
           MUTR3D~\cite{zhang2022mutr3d} & 6.0 \\
         \rowcolor{mygray}   \textbf{\name{}-S} (Ours) &  7.5 \\
           \bottomrule
    \end{tabularx}
    \label{tab:latency}
\end{table}

\subsection{Detection Results}
\subsubsection{Metrics.}
For the 3D detection task, we follow the official evaluation metrics from nuScenes~\cite{caesar2020nuscenes}, and report nuScenes Detection Score (NDS), mean Average Prediction (mAP), and five True Positive (TP) metrics including mean Average Translation Error (mATE), mean Average Scale Error (mASE), mean Average Orientation Error (mAOE), mean Average Velocity Error (mAVE), mean Average Attribute Error (mAAE). 
\input{latex/table/detection_val}

\input{latex/table/detection_test}

\mypara{Detection on nuScenes benchmark.}
As our \name{} can jointly optimize tracking and detection, we present the detection results on nuScenes test set and val set in \cref{tab:detection_test} and \cref{tab:detection_val} respectively, which demonstrate consistent improvements of our method in the detection task.
As a framework design for tracking, our model achieves comparable results ($62.7\%$ mAP and $68.0\%$ NDS in the test set) with the concurrent leading detection methods, e.g., HoP and Far3D, and outperforms previous end-to-end detection and tracking model Sparse4D-v3 by a significant margin of 5.7\% mAP and 2.4\% NDS in the test set.


\section{Additional Visualizations}
\subsection{Qualitative Results of PQI}
\input{latex/fig/bev}
We present qualitative results of our PQI module in \cref{fig:bev}. Our PQI module leverages learned certain priors, i.e., the predicted 2D object location and depth information to formulate initial queries. 
As shown in \cref{fig:bev}, the initial queries generated by our PQI module are accurately positioned within the regions of interest for the objects.
This indicates that our module effectively bootstraps latent query states of the objects, leading to improved object localization and tracking results.

\subsection{More Qualitative Results}
We provide additional qualitative results in \cref{fig:supp_vis}. We compare our \name{} with the previous state-of-the-art end-to-end tracker, PF-Track~\cite{pang2023PFtrack}, on various complex scenarios.
In challenging 3D MOT scenarios, characterized by multiple challenging factors such as occlusions and small target objects, \name{} successfully predicts a greater number of tracked bounding boxes with higher localization precision compared to PF-Track~\cite{pang2023PFtrack}.
Furthermore, the visualization of our attention scores demonstrates a higher concentration on the center of the objects, indicating that our model pays more attention to the target objects. This observation also highlights the effectiveness of our proposed UPD module with probabilistic attention in modeling the uncertainty associated with the object prediction.
\input{latex/fig/vis_supp}

\subsection{Video Demo} 
In addition to the figures, we have also attached a video demo in the supplementary materials, which consists of hundreds of tracking frames that provide a more comprehensive evaluation of our proposed approach.


%% file: latex/table/detection_val.tex
\begin{table*}[h]
  \caption{
  \textbf{Results of 3D detection on nuScenes \textit{val} dataset}.}
  \label{tab:detection_val}
  \centering
  \addtolength{\tabcolsep}{-1.5pt}
 \begin{tabularx}{1\textwidth}{l|c|cc|cccccc}
    \toprule
    Method & Backbone & mAP $\uparrow$ & NDS$\uparrow$ & mATE$\downarrow$ & mASE$\downarrow$ & mAOE $\downarrow$ & mAVE$\downarrow$ & mAAE $\downarrow$ \\

    \midrule
    \multicolumn{3}{l}{\textit{\textbf{Detection-only}}} \\
    \midrule
    
    PETR v2~\cite{liu2023petrv2} & R101 &  0.421& 0.524 &0.681 &0.267 &0.357 &0.377 &0.186 \\
    
    BEVDet4D~\cite{huang2022bevdet4d} & Swin-B & 0.426 & 0.552  &0.560 & 0.254  &0.317  &0.289 & 0.186  \\

    Cyclic~\cite{guo2024cyclic} & R101 & 0.433& 0.532 &0.639 &0.270 &  0.318 &0.416 &0.201\\
    
    BEVDepth~\cite{li2023bevdepth} & CNX-B & 0.462 & 0.558 & 0.540 &0.254 &0.353 & 0.379 & 0.200 \\
     AeDet~\cite{feng2023aedet} & CNX-B &  0.483&  0.581&  0.494&  0.261&  0.324 & 0.337& 0.195\\
    SOLOFusion~\cite{park2022time} & R101 & 0.483& 0.582 &0.503& 0.264 &0.381 &0.246& 0.207 \\
    StreamPETR~\cite{wang2023StreamPETR} & R101 & 0.504& 0.592& 0.569& 0.262& 0.315& 0.257& 0.199 \\
    
    HoP~\cite{zong2023temporal} & R101 & 0.454 &0.558 & 0.565& 0.265& 0.327 & 0.337 &0.194 \\
    Far3D~\cite{jiang2023far3d} & R101 & \bf0.510 & \bf0.594 & 0.551& 0.258 &0.372 &0.238 &0.195 \\

        \midrule \midrule
    \multicolumn{3}{l}{\textit{\textbf{Join Tracking and Detection}}} \\
    \midrule
    MUTR3D~\cite{zhang2022mutr3d} &  R101 &  0.349 & 0.434 & -- & -- & --   & -- & -- & \\
  DQTrack~\cite{li2023dqtrack} &V2-99 & 0.410 & 0.503 & -- & -- & -- &   -- & -- & \\
   PF-Track-F~\cite{pang2023PFtrack}  &  V2-99 & 0.399 &0.390& 0.727 & 0.268 &1.722 &0.887 &0.211 \\
   HSTrack~\cite{lin2024hstrack} & V2-99 & 0.418 & 0.510  & -- & -- & -- &   -- & -- &\\ 
    Sparse4D-v3~\cite{lin2023sparse4d} & R101 & 0.537 & 0.623 &0.511& 0.255& 0.306 &0.194 & 0.192\\
\rowcolor{mygray} \textbf{\name-F} (Ours) & ViT-L & \bf0.589 & \bf0.655 & 0.495 & 0.250 & 0.249 & 0.2100 & 0.1883 \\

    \bottomrule
    \multicolumn{6}{l}{\footnotesize{"\textbf{F}" represent on the full-resolution settings.} }
  \end{tabularx}

\end{table*}

%% file: latex/table/detection_test.tex
\begin{table*}[tb]
  \caption{
  \textbf{Results of 3D detection on nuScenes \textit{test} dataset}. $\dagger$ indicates the results obtained from our testing using the provided official model.}
  \label{tab:detection_test}
  \centering
  \addtolength{\tabcolsep}{-1.5pt}
 \begin{tabularx}{1\linewidth}{l|c|cc|cccccc}
    \toprule
    Method & Backbone & mAP $\uparrow$ & NDS$\uparrow$ & mATE$\downarrow$ & mASE$\downarrow$ & mAOE $\downarrow$ & mAVE$\downarrow$ & mAAE $\downarrow$ \\

    \midrule
    \multicolumn{3}{l}{\textit{\textbf{Detection-only}}} \\
    \midrule
    
    BEVDet4D~\cite{huang2022bevdet4d} & Swin-B & 0.451 & 0.569 & 0.511 & 0.241 & 0.386 & 0.301 & 0.121 \\
    Cyclic~\cite{guo2024cyclic} & R101 & 0.452& 0.549 &0.575& 0.255& 0.405 &0.407 &0.131 \\
    PETR v2~\cite{liu2023petrv2} & V2-99 & 0.506  & 0.592  &  0.536 &  0.243  & 0.359  & 0.349  & 0.120 \\
    BEVDepth~\cite{li2023bevdepth} & CNX-B & 0.520 & 0.609  & 0.445 & 0.243 & 0.352 & 0.347 & 0.127 \\
    BEVStereo~\cite{li2023bevstereo} & V2-99 & 0.525 & 0.610 & 0.431 & 0.246 & 0.358 & 0.357 & 0.138  \\
    SOLOFusion~\cite{park2022time} & CNX-B & 0.540 & 0.619 & 0.453 & 0.257 & 0.376 & 0.276 & 0.148 \\
    AeDet~\cite{feng2023aedet} & CNX-B & 0.531 & 0.620 & 0.439 & 0.247 & 0.344 & 0.292 & 0.130 \\
    StreamPETR~\cite{wang2023StreamPETR} & ViT-L & 0.620 & 0.676 & 0.470 & 0.241 & 0.258 & 0.236 & 0.134 \\
    
    HoP~\cite{zong2023temporal} & ViT-L & 0.624 & 0.685 & 0.367 & 0.249 & 0.353 & 0.171 & 0.131 \\
    Far3D~\cite{jiang2023far3d} & ViT-L & \bf0.635 & \bf0.687 & 0.432 & 0.237 & 0.278 & 0.227 & 0.130 \\

        \midrule \midrule
    \multicolumn{3}{l}{\textit{\textbf{Join Tracking and Detection}}} \\
    \midrule
   PF-Track-F$\dagger$~\cite{pang2023PFtrack}  &  V2-99 & 0.397 & 0.387 & 0.688 & 0.262 & 1.800 & 1.079 & 0.165 \\
    Sparse4D-v3~\cite{lin2023sparse4d} & V2-99 & 0.570 & 0.656 &0.412& 0.236& 0.312 &0.210 & 0.117\\
\rowcolor{mygray} \textbf{\name-F} (Ours) & ViT-L & \bf0.627 & \bf0.680 & 0.434 & 0.237 & 0.311 & 0.216 & 0.130 \\

    \bottomrule
    \multicolumn{9}{l}{\footnotesize{"\textbf{F}" represent on the full-resolution settings.} }
  \end{tabularx}
\end{table*}

%% file: latex/fig/bev.tex
\begin{figure*}[t]
  \centering
  \includegraphics[width=\linewidth]{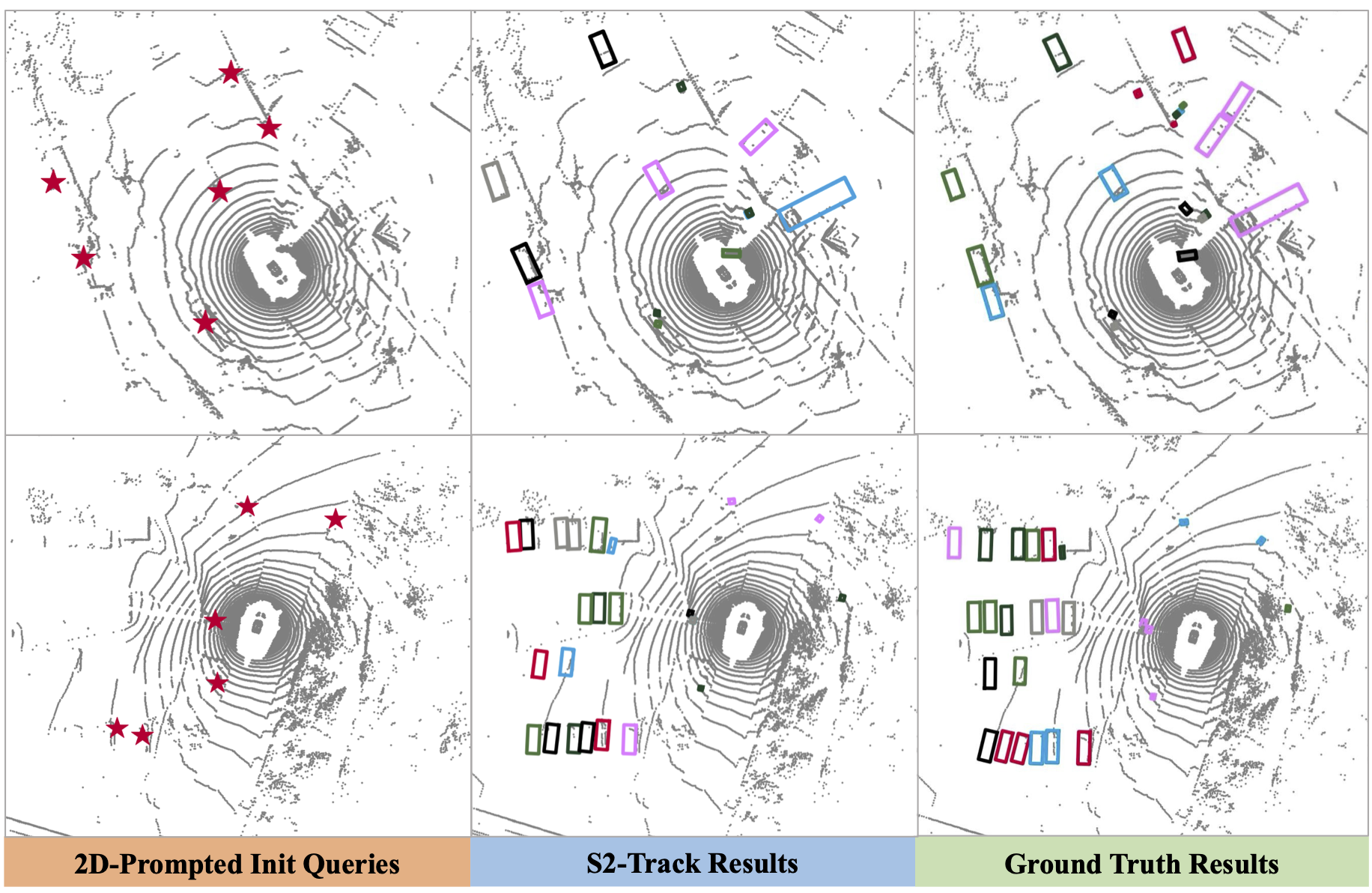}
  \caption{\textbf{Qualitative results of our \PQI{}}. The initial queries generated by our PQI module accurately locate the regions of interest for the objects, resulting in more accurate tracking results.
  }
  \label{fig:bev}
\end{figure*}

%% file: latex/fig/vis_supp.tex
\begin{figure*}[h!]
  \centering
  \includegraphics[width=\linewidth]{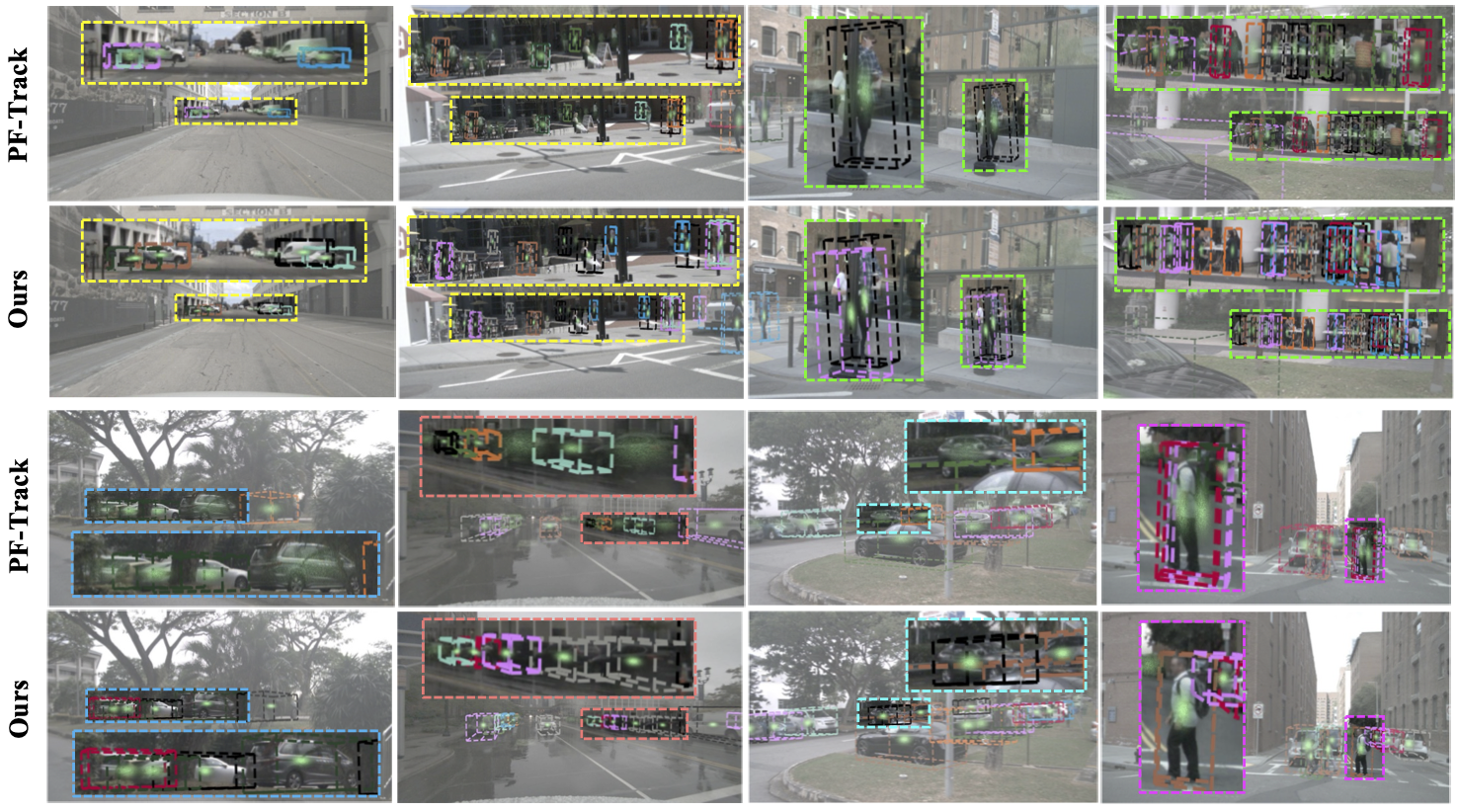}
  \caption{\textbf{More qualitative results on the nuScenes dataset}. The tracking results on several challenging tracking scenarios, including the small size of the target objects and the occlusions.
  Moreover, we plot the attention scores of object queries, which indicate how strongly the model focuses on the target objects. 
  A higher concentration of color represents a higher attention score and a stronger confidence in the corresponding object.
  }
  \label{fig:supp_vis}
\end{figure*}